\documentclass[runningheads, a4paper]{llncs}
\usepackage{graphicx}
\usepackage{adjustbox}

\usepackage{multirow}
\usepackage{cite}
\usepackage{amsmath}
\usepackage{amssymb}

\makeatletter
\newcommand{\printfnsymbol}[1]{%
  \textsuperscript{\@fnsymbol{#1}}%
}
\makeatother

\usepackage{hyperref}

\newcommand{\R}{\mathbb{R}}

\begin{document}
\title{Geometric Deep Learning for Post-Menstrual Age Prediction based on the Neonatal White Matter Cortical Surface
}
\titlerunning{Age Prediction based on the WM Surface}
%
\author{
Vitalis Vosylius\thanks{Equal contribution} \and
Andy Wang\printfnsymbol{1} \and
Cemlyn Waters\printfnsymbol{1} \and
Alexey Zakharov\printfnsymbol{1} \and
Francis Ward\printfnsymbol{1} \and
Loic Le Folgoc \and
John Cupitt \and
Antonios Makropoulos \and
Andreas Schuh \and
Daniel Rueckert \and
Amir Alansary
}


\authorrunning{Vosylius et al.}
%
\institute{Imperial College London, London, UK
}
\maketitle              
\begin{abstract}
Accurate estimation of the age in neonates is useful for measuring neurodevelopmental, medical, and growth outcomes.
In this paper, we propose a novel approach to predict the post-menstrual age (PA) at scan, using techniques from geometric deep learning, based on the neonatal white matter cortical surface. We utilize and compare multiple specialized neural network architectures that predict the age using different geometric representations of the cortical surface; we compare MeshCNN, Pointnet++, GraphCNN, and a volumetric benchmark. The dataset is part of the Developing Human Connectome Project (dHCP), and is a cohort of healthy and premature neonates. We evaluate our approach on $650$ subjects ($727$ scans) with PA ranging from $27$ to $45$ weeks. 
Our results show accurate prediction of the estimated PA, with mean error less than one week.

\end{abstract}
\section{Introduction} \label{sec:intro}
Precise age estimation in neonates helps measure the risk of neonatal pathology and organ maturity. Given that prematurity complications are the leading cause of all neonatal deaths, according to the world health organization (WHO)\footnote{https://www.who.int/news-room/fact-sheets/detail/preterm-birth}, precise age estimation may help to reduce the number of neonatal deaths significantly. 

There are different age terminologies during the prenatal period such as gestational age (GA), post-menstrual age (PA), and chronological age (CA) \cite{engle2004age}.
PA measures the time from the first day of
the last menstrual period and the birth time (GA) added to the time elapsed after birth (CA).
PA usually represents the age at scan taken during the
neonatal period after the day of birth, and is normally measured in weeks. 

The accuracy of the estimated PA is dependent on GA calculations; however, traditional methods of calculating the GA use the first day of the last menstrual period (LMP) as a reference point. As a result, the accuracy of these measurements is error-prone and relies on the patient's memory. Another method is to measure the diameter and circumference of the head, cranium, abdomen, and femur from 2D fetal ultrasound (US) images \cite{paladini2007sonographic}. However, this method also relies on operator expertise as well as the biological variations and inconsistencies in skull size approximation, which may lead to age approximation errors \cite{bottomley2009dating}. Therefore, developing automatic models that accurately predict the age can help with the diagnosis of several neurodevelopmental and psychiatric illnesses that are rooted in the early neonatal period \cite{rekik2016hybrid}. In this work, we propose a deep learning model that can accurately predict the PA using the white matter (WM) cortical brain surface. 

\textbf{Related Work:} 
A number of machine learning and statistical methods have been presented for perinatal brain age prediction based on brain image data or measurements. For example, Towes et al. \cite{toews2012feature} proposed a feature-based model for infant age prediction using scale-invariant image features extracted from T1-weighted MRI scans.
Brown et al. \cite{brown2017prediction} presented a method to predict the brain network age using random forests (RF) classification \cite{breiman2001random} from diffusion magnetic resonance imaging (dMRI) data. The output of their model was used to detect delayed maturation in structural connectomes for preterm infants.
Deprez et al. \cite{deprez2018segmentation} used logistic growth models to estimate the age of preterm infants based on segmented myelin-like signals in the thalami and brainstem.
Ouyang et al. \cite{ouyang2019differential} predicted the PA of preterm infants by measuring the temporal changes of cortical mean kurtosis (MK) and fractional anisotropy (FA) from non-Gaussian diffusion kurtosis imaging (DKI) and conventional diffusion tensor imaging (DTI).
Hu et al. \cite{hu2019hierarchical} predicted the infant age using a two-stage hierarchical regression model based on cortical features. 
Recently, Galdi et al. \cite{galdi2020neonatal} combined features from structural and diffusion MRI to model morphometric similarity networks (MSNs) that identify the inter-regional similarities between the features. The calculated MSNs were later used for predicting neonatal brain age. However, neonatal brain age prediction using deep learning methods has not yet been explored in the literature.

At the same time, other works have leveraged recent advances in deep learning for adult brain age prediction. For instance, Jiang et al. \cite{jiang2019predicting} presented a 3D convolutional neural network (CNN) to predict the brain age of healthy adults using structural network images as an input. 
Guti\'errez-Becker and Wachinger \cite{gutierrez2018deep} proposed a PointNet-based \cite{qi2017pointnet} architecture for predicting Alzheimer’s disease and brain age using multiple brain structures as an input for their model. 
Recently, Besson et al. \cite{besson2020geometric} utilize graph CNNs on a surface representation of the cortical ribbon for sex and age prediction on a data set of $6,410$ healthy subjects with ages ranging $6$-$89$ years. 
Compared to the previous methods, this work explores a number of directions for predicting the PA from the neonatal brain surface using geometric deep learning (GDL) approaches.

\textbf{Contributions:} To the best of our knowledge, this is the first work that evaluates age prediction from neonatal cortical surface representations. 
We evaluate a number of GDL architectures, namely PointNet++ \cite{qi2017pointnet++}, MeshCNN \cite{hanocka2019meshcnn}, and Graph CNN (GCN) \cite{kipf2016semi}.
Each architecture utilizes a different representation of the cortical surface: PointNet++ performs operations on a point cloud, MeshCNN operates on the edges of a mesh, and GCN operates directly on the adjacency, degree, and feature matrices of a graph. 
Our experiments show that these GDL techniques can accurately predict the PA from the neonatal cortical surface and outperform a 3D CNN benchmark that utilizes volumetric MRI data.
We use a large dataset of $650$ unique subjects ($727$ scans) with PA ranging from $26$ to $45$ weeks.

\textbf{Outline:} The rest of this paper is structured as follows: Section \ref{sec:background} provides a brief background to geometric deep learning (GDL). In Section \ref{sec:methods} we then describe the structural details of each GDL architecture that we use. Section \ref{sec:exp} outlines the details of our experiments. We finish by presenting our results followed by the conclusions and future directions. 

\section{Background} \label{sec:background}
The majority of work in deep learning for medical imaging typically focuses on the application of CNNs to Euclidean data, e.g. MRI and ultrasound images \cite{kamnitsas2017efficient, fetit2020deep}. 
However, CNN-based methods are usually restricted to exploit 2D or 3D volumetric images in Euclidean domains. 
This limits their application to complex geometric data defining embedded manifolds, e.g. brain cortical surfaces. In this case, convolutions are not well-defined, and the notion of CNN must be generalized to approximate functions in non-Euclidean domains. GDL methods aim to apply the power of CNNs to such non-Euclidean characterizations \cite{bronstein2017geometric, masci2015geodesic}. 
GDL methods in the literature can be categorized based on the representation of their input data as:
\begin{itemize}
    \item[$\bullet$] \textbf{Voxel based}: The nodes on a surface are projected to their corresponding (or nearest) locations in the 3D image \cite{wu20153d, brock2016generative}, where typical CNNs can be applied naturally. The main drawback is losing the surface representation, where two points far apart on the surface, in terms of its intrinsic geometry, can be very close in the volumetric or ambient Euclidean space. Furthermore, the projection to the 3D volume can introduce sampling accuracy errors. 

    \item[$\bullet$] \textbf{Point set based}: Models encode a set of points or nodes into 3D feature maps that can be processed by a typical neural network architecture. The best known models utilizing this approach are PointNet \cite{qi2017pointnet} and PointNet++ \cite{qi2017pointnet++}. They are agnostic to the origin of the point clouds they process and have an ability to leverage geometric features of non-Euclidean data. These models have been shown to achieve good performance efficiently; however, they cannot preserve relational information between the nodes on the surface. 

    \item[$\bullet$] \textbf{Graph based}: Models operate on graphs, which additionally encode connectivity information (edges) between the nodes on the surface. Variants of graph-based methods utilize GCNs to process the Laplacian of the graph in the spectral domain \cite{bruna2013spectral, henaff2015deep}. There are also special graph-based models that exploit meshes as their input graph. They are designed to operate on mesh edges and learn generalized convolution and pooling layers \cite{hanocka2019meshcnn}. The main drawback of these models is the difficulty to increase the model capacity or scale up for large input data.

\end{itemize}

\section{Methods} \label{sec:methods}
As mentioned previously, the attempt to generalize the power of CNNs to non-Euclidean data leads to a set of techniques known as geometric deep learning (GDL) \cite{bronstein2017geometric}. 
In this section, we present a number of GDL techniques for age-regression on brain surface representations: PointNet++ \cite{qi2017pointnet++}, MeshCNN \cite{hanocka2019meshcnn}, and GCN \cite{kipf2016semi}. 
The cortical surface meshes are extracted from MRI data as described in \cite{schuh2017deformable}. The point-cloud representation is extracted directly from the nodes (with node features), the graph from the same nodes together with the connectivity information, and the mesh representation is defined with geometric edge features described below.
We also implement a volumetric 3D CNN as a baseline. As noted in section \ref{sec:intro}, each architecture operates on a different representation of the brain surface, with each representation capturing subtly different geometric information. The architectures we present here also vary functionally, i.e., they perform different functions on the surface and therefore learn different abstractions of the brain surface. By proving that this range of GDL techniques performs brain-age regression with high accuracy, we show the utility of GDL to tasks related to the brain surface in general. We now describe the details of each architecture's structure and functional operation. 

\subsubsection{Voxel based:}
\begin{figure}[!htbp]
    \centering
    \includegraphics[width=\textwidth]{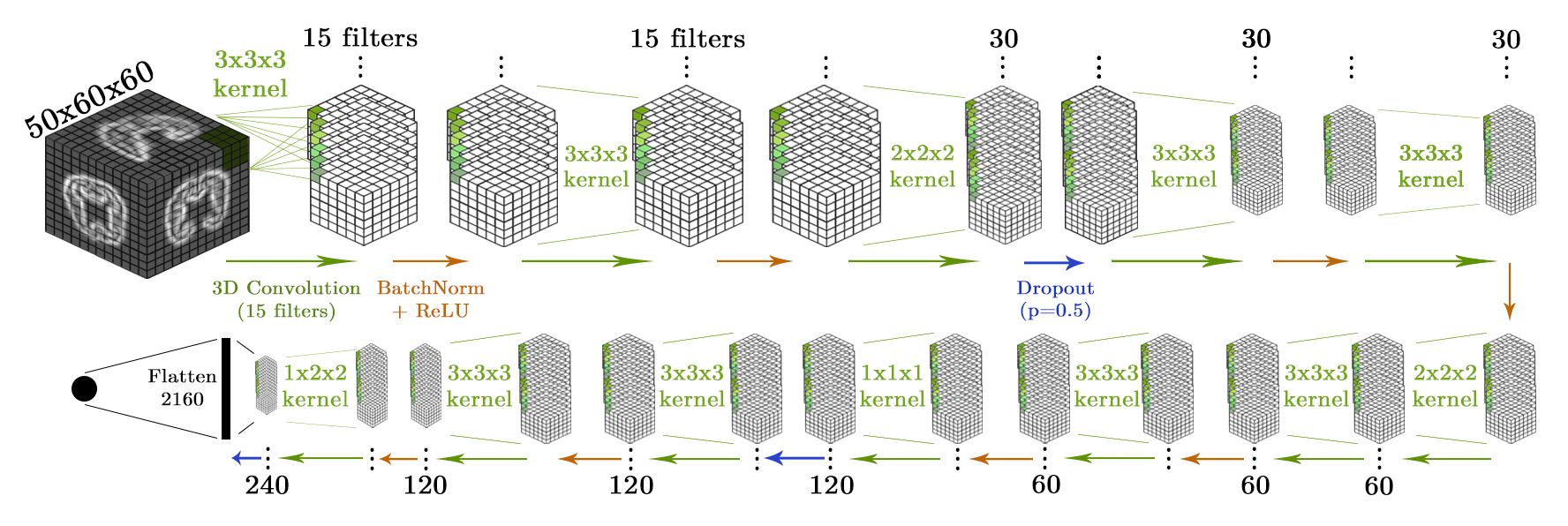}
    \caption{3D CNN architecture for age prediction using volumetric spatially-normalized gray-matter (GM) maps as an input.}
    \label{fig:volume3d}
\end{figure}

Similar to \cite{cole2017predicting}, we leverage a 3D CNN on spatially-normalized gray-matter (GM) maps as a baseline model. 
This baseline ensures the integrity of our experimentation and allows for a more in-depth analysis of the results using typical 3D volumetric images instead of surface representation.
For this approach, we consider a set of voxels, $V = (v^{(111)}, v^{(112)}, ..., v^{(XYZ)})$, where $(X, Y, Z)$ are the dimensions of the volumetric MRI image, and $v^{(xyz)} \in \R$ denotes voxel intensity at position $(x, y, z)$. 
The output of the $l$-th 3D convolutional layer for the $j$-th feature map at $(x, y, z)$ position is given by:
\begin{align}
    v^{(xyz)}_{(l+1)j} = \textrm{ReLU} \Big ( b_{(l)j} + \sum_m \sum^{A-1}_{a} \sum^{B-1}_{b} \sum^{C-1}_{c} W_{(l)jm}^{(abc)} \textrm{  }v^{(x+a)(y+b)(z+c)}_{(l)m} \Big ),
\end{align}
\noindent where $b_{(l)j}$ is the feature map bias term. $A$, $B$ and $C$ are kernel dimensions. $m$ indexes the set of feature maps in the $l$-th layer. $W_{(l)jm}^{(abc)}$ denotes weight value at kernel's position $(a, b, c)$ in $m$-th feature map. $v^{(x+a)(y+b)(z+c)}_{(l)m}$ is input value at $(x+a, y+b, z+c)$ in the $m$-th feature map. 
Combining several such layers with ReLU activation, dropout, 3D batch normalization, and the final linear layer, we are able to learn the weight matrices (kernels) optimizing for L1 loss function. 
Figure \ref{fig:volume3d} displays our proposed 3D CNN architecture.

\subsubsection{Point based:}
\begin{figure}[!ht]
    \centering
    \includegraphics[width=\textwidth]{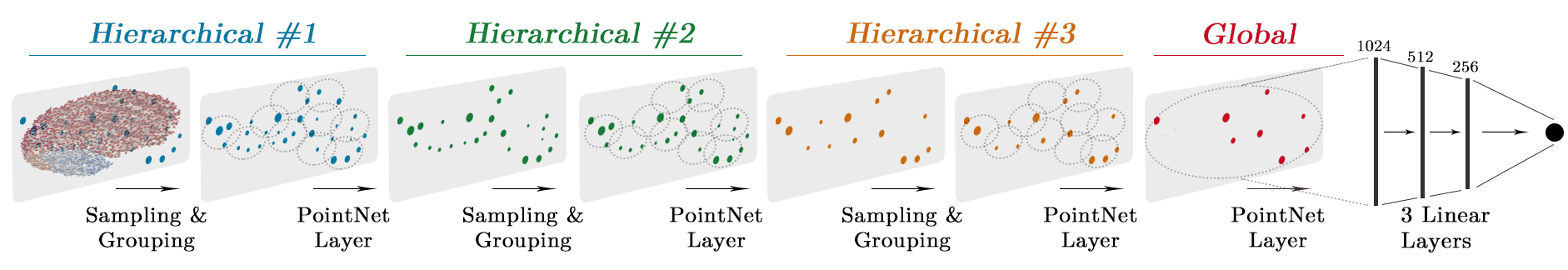}
    \caption{PointNet++ architecture for age prediction using the nodes on the brain surface as an input.}\label{fig:pointnet}
\end{figure}

We consider only the nodes on the brain surface, defined as a set of points (or point-cloud), $V = \{ v_1, v_2, ..., v_n \}$ with $v_i \in \R^d$ ($d=3$ in our case) and $n$ is number of nodes. A separate vector containing information about local features, $x_i \in \R^l$, is assigned to each point $v_i$, where $l$ is the number of local features considered. The original PointNet architecture \cite{qi2017pointnet}, is able to learn a function $f$ with the use of neural networks $\gamma$ and $h$ such that,
\begin{align}
    f(v_1, v_2, ..., v_n) = \gamma \big ( \textrm{max}_i \{ h(v_i) \} \big ).
\end{align}
This technique can approximate functions invariant to input permutation and linear transformations by using symmetric functions and alignment networks, respectively. 
PointNet++ \cite{qi2017pointnet++} extends this idea to hierarchical learning and includes sampling and grouping layers together with mini-PointNet layers. Sampling is performed using the farthest-point sampling (FPS), which provides better coverage than a completely random selection. Given the sampled centroids, the grouping layer then creates local point-sets around the centroids using a distance metric. PointNet++'s hierarchical structure allows for a progressive abstraction of the input yielding richer encoding of the global and local information described by the point-cloud. Figure \ref{fig:pointnet} shows a representation of the PointNet++ network architecture.

\subsubsection{Mesh based:}
\begin{figure}[!ht]
    \centering
    \includegraphics[width=\textwidth]{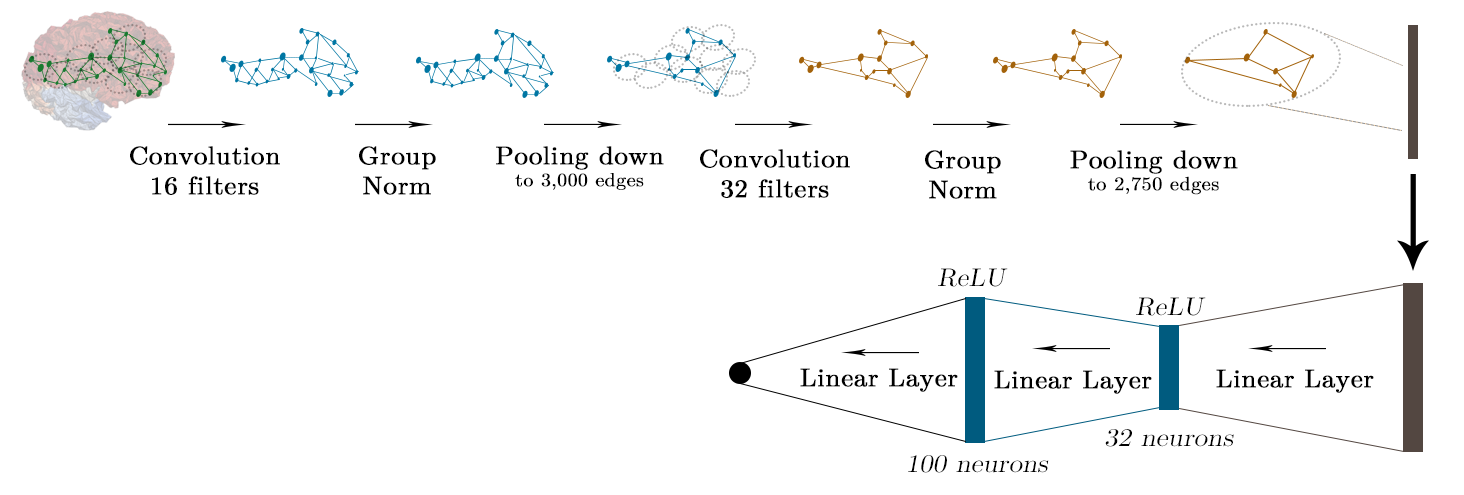}
    \caption{MeshCNN architecture for age prediction using the input brain mesh surface.}
    \label{fig:meshcnn}
\end{figure}

A mesh is defined as a pair of sets consisting of vertices ($V$) and connectivity information ($F$), with edges ($E$) defined as a set of connected pairs of vertices. In contrast to point-cloud techniques, mesh representation provides non-uniform, geodesic neighborhood information. Here, we use triangulated meshes as commonly used in the brain surface literature. 
The MeshCNN architecture \cite{hanocka2019meshcnn} consists of two main components for geometric learning: mesh convolution and pooling, see Fig. \ref{fig:meshcnn}. Both are operations defined over the input edges. Mesh convolution can be defined as:
\begin{align}
    e \cdot k_0 + \sum^4_{i=1} k_i \cdot e^i,
\end{align}
\noindent where $k$ is a kernel and $e$ is an edge feature. $e^i$ denotes an edge feature of the $i$-th neighboring edge, while total number of neighboring edges equal to $4$.
The input edge feature is a 5-dimensional vector containing geometric features: the dihedral angle, two inner angles, and two edge-length ratios for each face.
Importantly, symmetric functions are applied to ambiguous edge pairs to ensure invariance with respect to the permutation of the convolutional neighbors. 
The pooling component of MeshCNN uses the topology of the mesh to identify adjacency, and learn to non-uniformly collapse edges that contain the weakest features for the task at hand. Hence, it forms a process where the network exposes the important features while discarding the redundant ones. 
MeshCNN, to our knowledge, is the only architecture which exhibits such convolution and pooling properties specialized for triangulated meshes.

\subsubsection{Graph based:}
\begin{figure}[!ht]
    \centering
    \includegraphics[width=\textwidth]{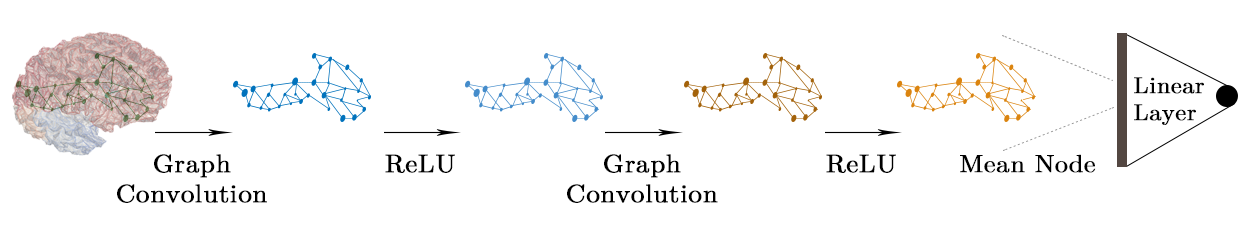}
    \caption{GCN architecture for age prediction using the input brain surface graph.}
    \label{fig:gcn}
\end{figure}

A graph, $\mathcal{G} = (V, E)$, is defined by a set of nodes ($V$) and a set of edges ($E$). Each node, $v_i$, represents a point on the brain surface and has an associated local feature vector, $X_i$. A graph convolution operation \cite{kipf2016semi} takes feature matrix $X^{(l)}$ in the $l$-th layer and outputs:

\begin{align}
    X^{(l+1)} = \sigma ( \hat{D}^{-\frac{1}{2}} \hat{A} \hat{D}^{-\frac{1}{2}} X^{(l)} W^{(l)}),
\end{align}

\noindent where $\hat{A} = A + I_N$ is the adjacency matrix of the graph modified by adding self-connections using identity matrix $I_N$. $N$ denotes the number of nodes on the graph. $\hat{D}=\sum_j \hat{A}_{ij}$ is the modified degree matrix. $X^{(l+1)}$ is the feature matrix in the ($l$+$1$)-th layer. 
Note, that the extracted local features of the surface are defined by a feature matrix at $l=0$ as $X^{(0)}$. Figure \ref{fig:gcn} demonstrates our proposed GCN architecture for age prediction.
The graph convolutional layers allow the network to learn meaningful feature vectors at each node using neighboring node information. 
Local features across the whole brain are aggregated by averaging feature vectors across all the nodes, which creates a global feature vector representing global geometric information of the graph. This global feature vector is used as input to a linear layer that outputs the predicted scan age.

Note the key similarities and differences between the previous architectures: The graph and mesh based methods operate on the connections between vertices, and local features are encoded as edge features, in contrast to point-wise features and operations on points. The graph methods therefore capture more information about the local geometric relations between points, which may require more computational resources to increase the capacity of the model (e.g. GPU memory). However, point-clouds are simple and unified structures that avoid the combinatorial irregularities and complexities of meshes, and thus it is easier to implement more efficient and larger PointNet-based models. 

Furthermore, the specialized graph and mesh convolutional operations leverage the intrinsic geodesic connections to learn hidden layer representations that encode both local graph structure and features of nodes. This is especially powerful when combined with MeshCNN's edge-pooling operations which expose and expand important features, whilst discarding irrelevant ones, allowing even richer encodings of the surface to be learned. This is comparable to PointNet++, in which the point sampling and grouping layers allow the hierarchical learning of informative points and encodes these into a rich high-level representation.

\section{Experiments} \label{sec:exp}
\subsection{Data}
The neonatal data used in this work are publicly available from the developing human connectome project (dHCP)\footnote{http://www.developingconnectome.org/second-data-release}. 
We excluded files with bad surface quality. 
Selected data consists of a cohort of $727$ total term and preterm neonatal MRI scans ($650$ unique subjects) with PA ranging from $27$ to $45$ weeks.
These data are split into $477$ $(65.7\%)$ train, $125$ $(17.15\%)$ test, and $125$ $(17.15\%)$ validation sets. 
To avoid any bias between between these subsets, the data split is stratified on multiple features: PA (scan age), birth age, and sex. 
The split is also done on the unique subjects to avoid data leakage from multiple scans between the subsets.
We validate our approach with experiments on the left and right hemispheres, as well as on both hemispheres merged. 
Surface files are decimated to 10,000 vertices to ensure comparable results between all proposed architectures. 
The surfaces are extracted from the segmented T2-weighted images in their native coordinates \cite{schuh2017deformable}.
Furthermore, we compare the performance of the GDL models using only geometric features, and with a range of other local point features, such as cortical thickness (CT), sulcal depth (SD), curvature (C), and the myelin map (MM). All cortical features are extracted using the dHCP structural MRI processing pipeline \cite{makropoulos2018developing}.

\subsection{Implementation}
\subsubsection{3D CNN:}
Similar to \cite{cole2017predicting}, we first segment the cortical gray matter from the 3D MRI scans.
Our 3D CNN consists of $12$ convolutional layers with ReLU activation and 3D batch normalization. Three dropout layers with $0.5$ probability are also added after every third convolutional layer. Setting the initial learning rate to $6.88\mathrm{e}$-$3$ , we train the model for $1000$ epochs and batch size $32$, with Adam optimizer and a scheduler, which decays the learning rate by a factor of $\gamma = 0.9795$ after every epoch. The input images are down-sampled and smoothed using a discrete Gaussian kernel of size 8.
\subsubsection{PointNet:}
For the implementation of PointNet++, we employ the PyTorch Geometric\footnote{https://github.com/rusty1s/pytorch\_geometric} python library. We use three hierarchical levels containing a sampling layer, a grouping layer and a PointNet layer. We also use ReLU activation functions and batch normalization \cite{batchnorm}. After hierarchical levels, another PointNet layer and global max pooling produce the final vector of size $1024$ which is input to $3$ fully connected layers producing the final age prediction. Mean square error (MSE) loss criterion, Adam optimizer, and a learning rate scheduler are employed to train the model, with an initial learning rate of $1\mathrm{e}$-$3$ and a lower bound of $5\mathrm{e}$-$5$.
\subsubsection{MeshCNN:}
We adapt the original MeshCNN code\footnote{https://github.com/ranahanocka/MeshCNN} for the task of age regression. We use group normalization with two groups, an MSE loss criterion, Adam optimizer, and a ReduceOnPlateau learning rate scheduler, with an initial learning rate of $3\mathrm{e}$-$4$ and a lower bound of $3\mathrm{e}$-$5$. Because of the expensive GPU memory requirements of the MeshCNN implementation we use a mini-batch of size one.
Weights are initialized using Kaiming normal initialization \cite{he2015delving}.
\subsubsection{GCN:}
Our GCN implmentation is based on the DGL\footnote{https://www.dgl.ai} library using the graph convolutions (GCs) defined in \cite{kipf2016semi}. The network architecture consists of two GC layers, each using ReLU activation, finally the mean feature vector is calculated across all nodes in the graph before being fed to the linear layer. The weights of the graph convolutional layers are initialized using Glorot uniform \cite{glorot2010understanding} and the biases were set to zero. We use Adam optimizer with a cosine annealing learning rate, starting at $8\mathrm{e}$-$4$ and decreasing down to $1\mathrm{e}$-$6$ with $T_{\textrm{max}}$ set to $10$. Our implementation of all previous architectures is publicly available on GitHub: \hyperlink{https://github.com/andwang1/BrainSurfaceTK}{https://github.com/andwang1/BrainSurfaceTK}.
%
\section{Results}
Table \ref{results} shows the results of PA prediction using the proposed architectures. 
With each of our GDL models, we report mean absolute error (MAE) less than one week on both the validation and test sets. 
The best performance (MAE$_{val} = 0.701$, MAE$_{test} = 0.6211$ weeks) is attained by PointNet++ with added (local) cortical features (cortical thickness, curvature, and sulcal depth). 
This is competitive with the 3D CNN benchmark and indeed, performs better on the test set ($0.62$ against $0.82$ weeks). The variation in performance between the validation and test sets is due to slight differences in feature distributions between these sets, since the number of samples is relatively small and there are many constraints to satisfy in the splits (sex and age distribution etc).
Due to containing local connections between points, the data in mesh form is expected to carry more information than the point cloud representation. 
Despite this, the results show that PointNet++ outperforms MeshCNN in both validation and test MAE. 
However, an important observation to note is that the MeshCNN implementation used to generate these results uses only $8$k parameters compared to PointNet's $1.5$M. 
This suggests that the data representation and the features used by MeshCNN are very informative and suitable for the regression tasks. 
On the other hand, our GCN implementation had only $68$k trainable parameters. 
\begin{table}[!htb]
\centering
\resizebox{\textwidth}{!}{
\begin{tabular}{c|c|c|c|c|c}
\hline
\textbf{Model} &
  \textbf{Hemisphere} &
  \textbf{Input Size} &
  \textbf{Validation Error (wks)} &
  \textbf{Test Error (wks)} &
  \textbf{Cortical Features} \\ \hline
\textbf{3D CNN}                      & GM Maps & 50x60x60   & $0.6765 \pm 0.5821$  & $0.8221 \pm 0.6858$    & -         \\ \hline
\multirow{3}{*}{\textbf{PointNet++}} & Right   & 10k & $0.8980 \pm 0.6651$         & $1.0417 \pm 0.9201$    & -         \\
                                     & Left    & 10k & $0.9380 \pm 0.7012$         & $0.9810 \pm 0.9043$    & -         \\
                                     & Whole   & 10k & $0.8128 \pm 0.7513$         & $0.8100 \pm 0.6918$    & -         \\ \hline
\multirow{3}{*}{\textbf{\begin{tabular}[c]{@{}c@{}}PointNet++\\ (with cortical features)\end{tabular}}} 
                                     & Right   & 10k & $0.7217 \pm 0.6138$         & $0.7084 \pm 0.5982$    &  CT, C, SD \\
                                     & Left    & 10k & $0.8140 \pm 0.5813$         & $0.6915 \pm 0.6647$    & CT, C, SD \\
                                     & Whole   & 10k & $\textbf{0.7010} \pm \textbf{0.6209}$ & $\textbf{0.6211} \pm \textbf{0.4784}$ & CT, C, SD \\ \hline
\multirow{3}{*}{\textbf{MeshCNN}}    & Right   & 10k & $0.8273 \pm 0.6692$         & $0.8797 \pm 0.6691$    & -         \\
                                     & Left    & 10k & $0.8986 \pm 0.6590$         & $0.8811 \pm 0.7056$    & -         \\
                                     & Whole   & 10k & $0.8810 \pm 0.6746$         & $0.9555 \pm 0.6513$    & -         \\ \hline
\multirow{3}{*}{\textbf{GCN}}        & Right   & 10k & $1.3029 \pm 1.0266$         & $1.3391 \pm 1.0307$    & -         \\
                                     & Left    & 10k & $1.2455 \pm 0.9432$         & $1.2455 \pm 0.9804$    & -         \\
                                     & Whole   & 10k & $1.1208 \pm 1.1208$         & $1.1617 \pm 0.9348$    & -         \\ \hline
\multirow{3}{*}{\textbf{\begin{tabular}[c]{@{}c@{}}GCN\\ (with cortical features)\end{tabular}}} 
                                     & Right   & 10k & $0.7956 \pm 0.9819$         & $0.7793 \pm 0.6818$    & CT, C, SD \\
                                     & Left    & 10k & $0.7589 \pm 0.6395$         & $0.7273 \pm 0.6403$    & CT, C, SD \\
                                     & Whole   & 10k & $0.7511 \pm 0.6205$         & $0.7182 \pm 0.5741$    & CT, C, SD \\ \hline
\end{tabular}}
\caption{The detailed results from the proposed architectures using: left or right hemispheres, or the whole brain. 
}
\label{results}
\end{table} 

Table \ref{results2} shows a summary of previous reported results for age prediction in the literature.
Although our results are not directly comparable with the reported works in the table, because of the differences in the employed input data modalities, validation techniques and variations in age ranges, our prediction error is the lowest. We also use the biggest dataset size for our experiments compared to the other published works.
\begin{table}[!htb]
\resizebox{\textwidth}{!}{%
\begin{tabular}{l|l|l|c|c}
\hline
\multicolumn{1}{c|}{\textbf{Method}} & \multicolumn{1}{c|}{\textbf{Input}} & \multicolumn{1}{c|}{\textbf{Size of the Data}} & \textbf{Age Range (wks)} & \textbf{Error (wks)} \\ 
\hline
Toews et al. \cite{toews2012feature}        & Scale-invariant T1w features.  & 92 subjects (230 infant structural MRIs) & 1.1–84.2 CA & $10.28$ \\
Brown et al. \cite{brown2017prediction}     & FA-weighted structural connectivity  & 168 DTIs                & 27–45 PA     & $1.6$             \\
Ouyang et al. \cite{ouyang2019differential} & Cortical FA and MK (mean kurtosis)   & 89 preterm infants      & 31.5–41.7 PA & $1.41$            \\
Deprez et al. \cite{deprez2018segmentation} & Signals in the thalami and brainstem & 114 preterm infants     & 29–44 PA     & $2.56$            \\
Hu et al. \cite{hu2019hierarchical}         & Cortical measures      & 50 healthy subjects (251 longitudinal MRIs) & 2–6.9 CA & $1.58 \pm 0.04$ \\
Galdi et al. \cite{galdi2020neonatal}       & Structural and diffusion MRI & 105 neonates (59 preterm and 46 term) & 38–44.56 PA & $0.70 \pm 0.56$ \\
PointNet++ (proposed)                       & WM surface nodes                     & 650 subjects (727 MRIs) & 27–45 PA     & $\textbf{0.6211} \pm \textbf{0.4784}$ \\
MeshCNN (proposed)                          & WM surface mesh                      & 650 subjects (727 MRIs) & 27–45 PA     & $0.9555 \pm 0.6513$           \\
GCN (proposed)                              & WM surface graph                     & 650 subjects (727 MRIs) & 27–45 PA     & $0.7182 \pm 0.5741$          \\ 
\hline        
\end{tabular}}
\caption{Results from previous works for age prediction. Our experiments utilize the biggest dataset for evaluation, showing the lowest error.}
\label{results2}
\end{table}

\section{Conclusion and Discussions}

To the best of our knowledge, this work presents the first study to assess a number of geometric deep learning (GDL) architectures on the task of PA regression based on the neonatal white matter surface. 
We compare several GDL architectures from the literature that utilize different representations of the brain surface, either point-clouds, meshes, or graphs. 
We compare our models against a 3D CNN baseline architecture for age prediction using the 3D volumetric gray matter maps. 
Models are evaluated on a large cohort of $727$ term and preterm scans ($650$ subjects) with a wide PA range of $27-45$ weeks.
Our results show accurate prediction of the estimated PA, with the best model's average error around $0.62$ weeks. 
It is the lowest error compared to previously published works for predicting PA. 

\textbf{Limitation and Future Direction:} 
We note that there is a trade-off between graph and point based methods such that graph representations capture more geometric information and the networks are more efficient (in that they attain similar performance with much fewer parameters). However, graph based techniques are also computationally expensive which may limit the model size. On the other hand, PointNet methods are computationally efficient but the points do not capture as much information, so much larger models are needed.
Similar to typical CNNs, the proposed GDL architectures can be sensitive to the input data, e.g. errors on the extracted surface. Hence, as well as suggesting developmental abnormalities, surfaces for which our models predict anomalously inaccurate PA may have been extracted incorrectly.
A future direction of our work will be to investigate the application of GDL to the association of brain regions with accurate age prediction.
GDL could also be applied to the classification of preterm neonates from the brain surface, and may provide insights into the development of the neonatal cortical surface.

\bibliographystyle{splncs04}
\bibliography{references}

\end{document}